%% file: main.tex
\documentclass[10pt,twocolumn,letterpaper]{article}

\usepackage{iccv}              
\usepackage{multirow, soul}
\usepackage{placeins}

\usepackage{xcolor, kotex}
\usepackage{color,colortbl}


\definecolor{iccvblue}{rgb}{0.21,0.49,0.74}
\usepackage[pagebackref,breaklinks,colorlinks,allcolors=iccvblue]{hyperref}
\usepackage[accsupp]{axessibility}

\def\paperID{13419} 
\def\confName{ICCV}
\def\confYear{2025}

\newcommand\blfootnote[1]{%
  \begingroup
  \renewcommand\thefootnote{}\footnote{#1}%
  \addtocounter{footnote}{-1}%
  \endgroup
}

\title{DIA: The Adversarial Exposure of Deterministic Inversion in Diffusion Models}

\author{Seunghoo Hong$^{1, \dagger}$ ~ Geonho Son$^{2, \dagger}$ ~ Juhun Lee$^{1}$ ~ Simon S. Woo$^{1,2,\star}$\\
$^{1}$Dept. of Artificial Intelligence, $^{2}$Dept. of Computer Science \& Engineering \\
Sungkyunkwan University, South Korea\\
{\tt\small \{hoo0681, sohn1029, josejhlee, swoo\}@g.skku.edu}
}

\DeclareMathOperator*{\argmax}{arg\,max}

\begin{document}
\maketitle
\input{sec/0_Abstract}

\input{sec/1_Introduction}

\input{sec/2_RelatedWork}
\input{sec/3_Preliminary}
\input{sec/4_Method}
\input{sec/5_Experiment}

\input{sec/6_Discussion}

\input{sec/7_Conclusion}

{
    \small
    \bibliographystyle{ieeenat_fullname}
    \bibliography{main}
}

\clearpage

\appendix

\input{sec/99_Appendix}
\end{document}

%% file: sec/0_Abstract.tex
\begin{abstract}
Diffusion models have shown to be strong representation learners, showcasing state-of-the-art performance across multiple domains. Aside from accelerated sampling, DDIM also enables the inversion of real images back to their latent codes. A direct inheriting application of this inversion operation is real image editing, where the inversion yields latent trajectories to be utilized during the synthesis of the edited image. Unfortunately, this practical tool has enabled malicious users to freely synthesize misinformative or deepfake contents with greater ease, which promotes the spread of unethical and abusive, as well as privacy-, and copyright-infringing contents. While defensive algorithms such as AdvDM and Photoguard have been shown to disrupt the diffusion process on these images, the misalignment between their objectives and the iterative denoising trajectory at test time results in weak disruptive performance. In this work, we present the \textbf{D}DIM \textbf{I}nversion \textbf{A}ttack (DIA) that attacks the integrated DDIM trajectory path. Our results support the effective disruption, surpassing previous defensive methods across various editing methods. We believe that our frameworks and results can provide practical defense methods against the malicious use of AI for both the industry and the research community. Our code is available here: \url{https://anonymous.4open.science/r/DIA-13419/}.

\blfootnote{
$^\dagger$ Co-authors with equal contributions} 
\blfootnote{
$^\star$ Corresponding author
}
\end{abstract}

%% file: sec/1_Introduction.tex
\section{Introduction}
Diffusion models have reshaped the way we generate images over the last few years~\cite{ho2020denoising,song2020score}. The artifact of such impact is both evident in the compounding empirical evidence of their performance paired with solid rooting from thermodynamics~\cite{sohl2015deep}. The joint acceleration in diffusion research and high-quality dataset curation catalyzed the emergence of large-scale text-to-image latent diffusion models such as Stable Diffusion. Equipped with classifier-free guidance, Stable Diffusion has shown extrapolative image generation capability for unseen and complex prompting~\cite{schuhmann2022laion,rombach2022high}. 
\begin{figure}[t]
\centering
\includegraphics[width=\columnwidth]{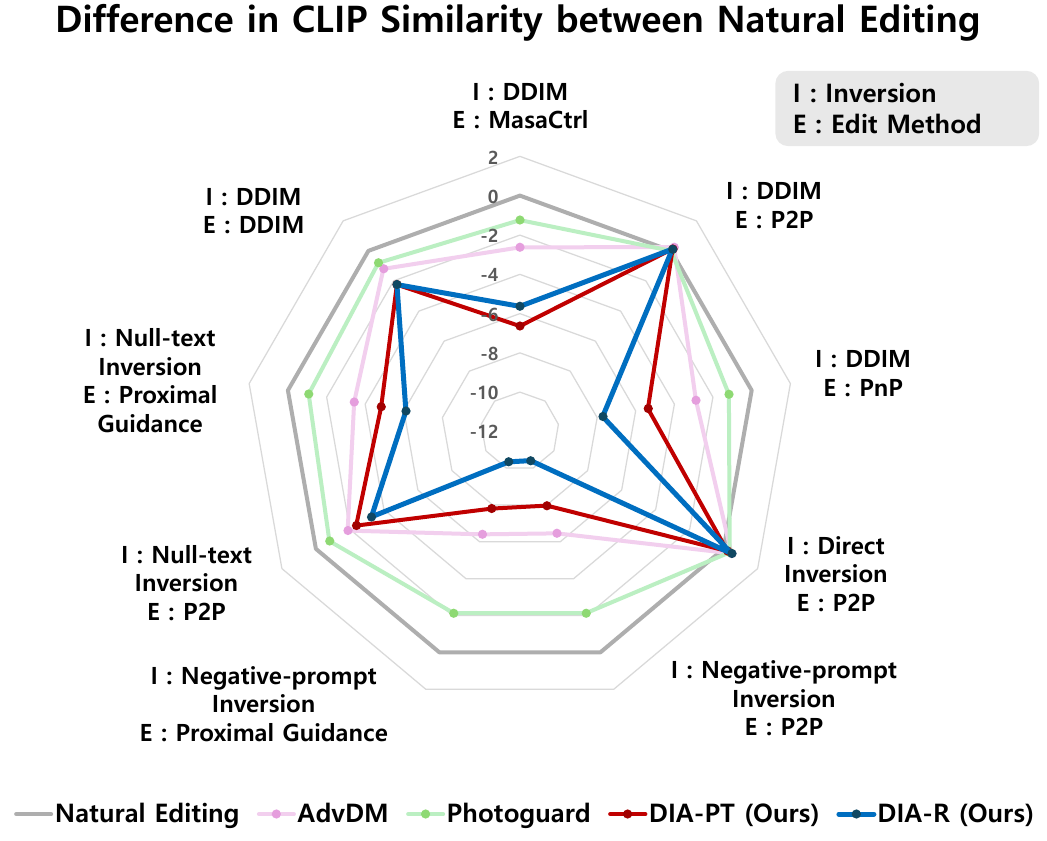}
\caption{\textbf{CLIP similarity score difference between Natural Editing and disruption methods.} Our methods DIA-PT/R demonstrate good semantic disruption performance across various combinations of Inversions and Edits. Lower scores equate to stronger disruption.}
\label{fig:1page}
\vspace{-8pt}
\end{figure}

As research continues to enhance the generalization and fidelity of diffusion-based models in representing data distributions $p(x)$, various downstream applications, including real image editing, have benefited from these advances~\cite{hudson2023soda}. An adjacent, major application is real image editing. Indeed, real image editing is a long-standing task, where its fundamental goal is to modify the given image according to an arbitrary editing instruction. While the success of an edit can be attributed to multiple key factors such as disentanglement, edit locality, and cohesiveness~\cite{ju2023direct}, identity preservation is by far the most indispensable. Thankfully, through the deterministic forward diffusing of Denoising Diffusion Implicit Models (DDIM) inversion~\cite{song2020denoising,dhariwal2021diffusion}, one can gain access to a corresponding latent code of the real image, which allows not only a basic reconstruction of the original image, but also serves as an initial latent code at which the editing offset is applied.
Inspired by this elementary approach, many methods have built upon it to further enhance editability, introducing new paradigms~\cite{cao2023masactrl,hertz2022prompt,mokady2023null,ju2023direct}. These methods have had great implications in the multimedia domain due to the model's stability and zero-shot generalization performance.

However, despite the inherent benefits of open-source technologies, malicious users can exploit them to synthesize and create fake contents that poses a serious privacy threat and unethical contents, impacting not only individuals but also our society by enabling the creation of ``Not safe for work'' (NSFW)~\cite{nsfw}, and ``Child Sexual Exploitation Material'' (CSEM) contents~\cite{csem}, as well as evidence manipulation. AdvDM and Photoguard~\cite{liang2023adversarial,salman2023raising} are pioneering works that motivated disrupting the diffusion process through adversarial perturbations as defensive methods against the misuse of real image editing. While AdvDM misleads the U-Net from denoising correctly, Photoguard attacks the image encoder, so that the diffusion process starts from a disrupted latent image. However, their effectiveness against inversion-based editing methods is significantly less pronounced as previous works have pointed out that the extent of their disruption is sub-performant or not protective enough against most editing approaches~\cite{salman2023raising,wang2023simac,cao2024impress}. In essence, the introduction of adversarial deviation from previous immunization methods does not take into account the recursive diffusion chain process, which inhibits them from eliciting digression away from the original image. 


In this work, we propose \textbf{D}DIM \textbf{I}nversion \textbf{A}ttack (DIA) which demonstrates significant immunization effectiveness against inversion-based editing methods. In this framework, we explore trajectory-based approaches for disrupting the deterministic inversion process by leveraging diffusion models.

By reformulating the inversion process, we identify that we can couple the diffusion process trajectory with different plausible objectives. Namely, we investigate 1) DIA-PT, where the objective targets the inversion \textbf{P}rocess \textbf{T}rajectory, and 2) DIA-R, which utilizes the \textbf{R}econstruction loss after encoding and decoding the image. Furthermore, the objective targeting the trajectory relaxes GPU's VRAM memory constraints thanks to decomposing the back-propagation through the Vector-Jacobian Product. Our findings show that the loss function that directly negates the learned diffusion trajectory shows the most optimal performance. To the best of our knowledge, there is no existing work centered around disrupting inversion-based real-image editing methods. 

In summary, our contributions are summarized below:
\begin{itemize}

 \item We first identify that current disruption methods lack an objective alignment with every inversion-based editing method. Therefore, we propose methods that disrupt diffusion trajectories in the DDIM Inversion process to prevent malicious image-editing.

 \item 
 We extend the differentiable DDIM trajectory in a memory-friendly manner using decomposed back-propagation and Vector-Jacobian product. Under this overarching setup of DDIM Inversion Attack (DIA), we propose a \textit{practical} attack that focuses on isolating and attacking the DDIM process trajectory based on the inversion formulation.

 \item We conducted extensive disruption experiments on PIE-Bench, which covers a diverse range of images and scenarios for editing methods. These experiments demonstrate that our proposed method achieves state-of-the-art disruption and works effectively across various DDIM Inversion-based editing methods. 
\end{itemize}

%% file: sec/2_RelatedWork.tex
\section{Related Works}
\subsection{Diffusion Models}
Diffusion models are trained to model the backward diffusion process (denoising) by matching it to the forward diffusion process (noising)~\cite{ho2020denoising}. Once fully trained, the model can generate clean image $x_0$ by an iterative backward process departing from $x_T \sim \mathcal{N}(0,~1)$. In the forward diffusion process, one can yield data $x_t$ from time step $t$:
\begin{equation}
\begin{aligned}
q\left(x_t \mid x_{t-1}\right) & := \mathcal{N}\left(x_t ; \sqrt{\alpha_t} x_{t-1},\left(1-\alpha_t\right) \mathbf{I}\right), \\
\end{aligned}
\end{equation}

\noindent where $\alpha_t$ follows a pre-defined scheduling with $t$ ranging from $1$ to $T$. 
Moreover, we can obtain $x_t$ directly from $q(x_t|x_0)$ analytically. Then, diffusion training consists of matching the prediction $p_\theta(x_t|x_{t+1})$ with $q(x_t|x_0)$. 
In latent diffusion models, the image encoder $\mathcal{E}$ maps image $x$ to its latent code $z$ and the diffusion occurs in the latent space ~\cite{rombach2022high}. Additionally, text embedding $c$ is fed as a conditional signal to the model. Then, the standard diffusion loss is augmented as follows:
\begin{equation} 
\mathcal{L}_\text{LDM}=\mathbb{E}_{\mathcal{E}(x), t, c, \epsilon \sim \mathcal{N}(0,1)}\left[\left\|\epsilon-\epsilon_\theta\left(z_t, c, t\right)\right\|_2^2\right].
\end{equation}
\subsection{DDIM Sampling \& Inversion}
As a Markovian process, Denoising Diffusion Probabilistic Models (DDPM) ~\cite{ho2020denoising} add random noise at each sampling step of the diffusion backward process $q(x_{t-1} | x_t)$. Thus, the sampling of $x_0$ requires close to $T$ steps. To accelerate sampling, DDIM~\cite{song2020denoising} introduces a non-Markovian forward process. Namely, it forgoes the introduction of random noises during sampling, making the process effectively deterministic:
\begin{align} \label{eq:ddim_sampling}
x_{t-1} &= \sqrt{\bar{\alpha}_{t-1}}\underset{\text{predicted }x_0}{\left( \frac{x_t - \sqrt{1-\bar{\alpha}_t}\epsilon_\theta(x_t,t)}{\sqrt{\bar{\alpha}_t}} \right)} + \\ \nonumber 
&\underset{\text{direction pointing to }x_t}{\sqrt{1-\bar{\alpha}_{t-1} - \sigma_t^2}\epsilon_\theta(x_t,t)} + \underset{\text{random noise}}{\sigma_t z},~\\ \nonumber
&z \sim \mathcal{N}(0, I).
\end{align}
Intuitively, DDPM sampling is equivalent to the stochastic differential equation (SDE) process~\cite{song2020score}. With the removal of the stochastic variable, DDIM sampling corresponds to the ordinary differential equation (ODE) process. Note that we can control the introduction of randomness by adjusting $\sigma_t$, where $\sigma_t = 0$ makes the sampling deterministic. Analogous to solving regular ODEs,~\cite{dhariwal2021diffusion} showed that one can reverse the travel, going from data to noise:
\begin{equation} \label{eq:ddim_inversion}
x_{t}=\frac{\sqrt{\bar{\alpha}_{t}}}{\sqrt{\bar{\alpha}_{t-1}}}x_{t-1}+\sqrt{\bar{\alpha}_{t}}(\lambda(t-1))\epsilon(x_t,t),  
\end{equation}
where $\lambda(t) := \sqrt{\frac{1}{\bar{\alpha}_{t+1}}-1} -\sqrt{\frac{1}{\bar{\alpha}_{t}}-1}$. This effectively represents we can obtain the latent $x_T$ corresponding to any arbitrary input data by sampling with Eq.~\ref{eq:ddim_inversion}. Then, denoising from the inverted latent $x_T$ leads back to the original image. 
One should note that Eq.~\ref{eq:ddim_sampling} utilizes a linearization assumption to use $\epsilon(x_{t-1},t)$ to approximate the non-existent $\epsilon(x_{t},t)$.
 
\subsection{DDIM Inversion-based Real Image Editing}

Another significant area of interest is the coherent modulation of real images. Within the diffusion framework, SDEdit is one of the first image-to-image methods~\cite{meng2021sdedit}. It relies on stochastic forward diffusion to first noise the image to a certain point and denoise back to image space with the desired text condition.  To overcome the stochasticity presented in SDEdit, inversion methods, notably DDIM inversion, allow mapping an image back to its corresponding latent code in a deterministic fashion. However, due to the substantial reconstruction error and low editability of DDIM inversion, several research works focused on improving those. Namely, Null-text Inversion~\cite{mokady2023null} minimizes the distance between the inversion and the reconstruction trajectory by optimizing the unconditional text embedding. In Negative-Prompt Inversion~\cite{miyake2023negative}, the unconditional text embedding during reconstruction is set to be the conditional text embedding used during inversion. PnP inversion~\cite{ju2023direct} rectifies the source by introducing an offset distance between the inversion and reconstruction latents, while maximizing the editability of the target diffusion branch. Another approach, P2P, injects the modulated attention maps into the reverse diffusion~\cite{hertz2022prompt}. MasaCtrl extracts saliency maps from cross-attention maps from both the forward and reverse diffusion to build a masked-guided mutual self-attention operation~\cite{cao2023masactrl}.

\subsection{Real Image Protection}
Two major uses of real images with diffusion are image editing and reference-based model personalization/stylization. Similarly, protection methods for both applications have been proposed. 

Notably, Photoguard~\cite{salman2023raising}and PID~\cite{li2024pid} propose to attack the image encoders in LDMs such that the diffusion processes diffuses a disrupted image latent. Similarly, instead of yielding adversarial gradient with the target image encoder, Glaze~\cite{shan2023glaze} relies on a off-the-shelf feature extractor as a surrogate image encoder.




Another class of attacks, first introduced by AdvDM~\cite{liang2023adversarial}, focuses on attacking the diffusion process directly by maximizing the diffusion loss. Thus, in a broader scope, the diffusion and image encoder loss set the paradigm for the following works, where some of them apply these update ``engines" to protect against unconsented model personalization~\cite{ahn2024imperceptible, van2023anti, wang2023simac,zheng2023understanding}. To overcome both time and memory complexity and robustness of the protective noise, Score Distillation Sampling (SDS) based approach relies on approximating the Jacobian as $\mathcal{J}_{z_t}\epsilon_{\theta}(z_t,t) \approx  I$ to bypass backward computation and, counterintuitively, minimizes semantic loss. Yet, so far, no method has been proposed to combat directly against unconsented inversion-based image editing.

%% file: sec/4_Method.tex
\begin{figure*}[t]
\centering
\includegraphics[width=\textwidth]{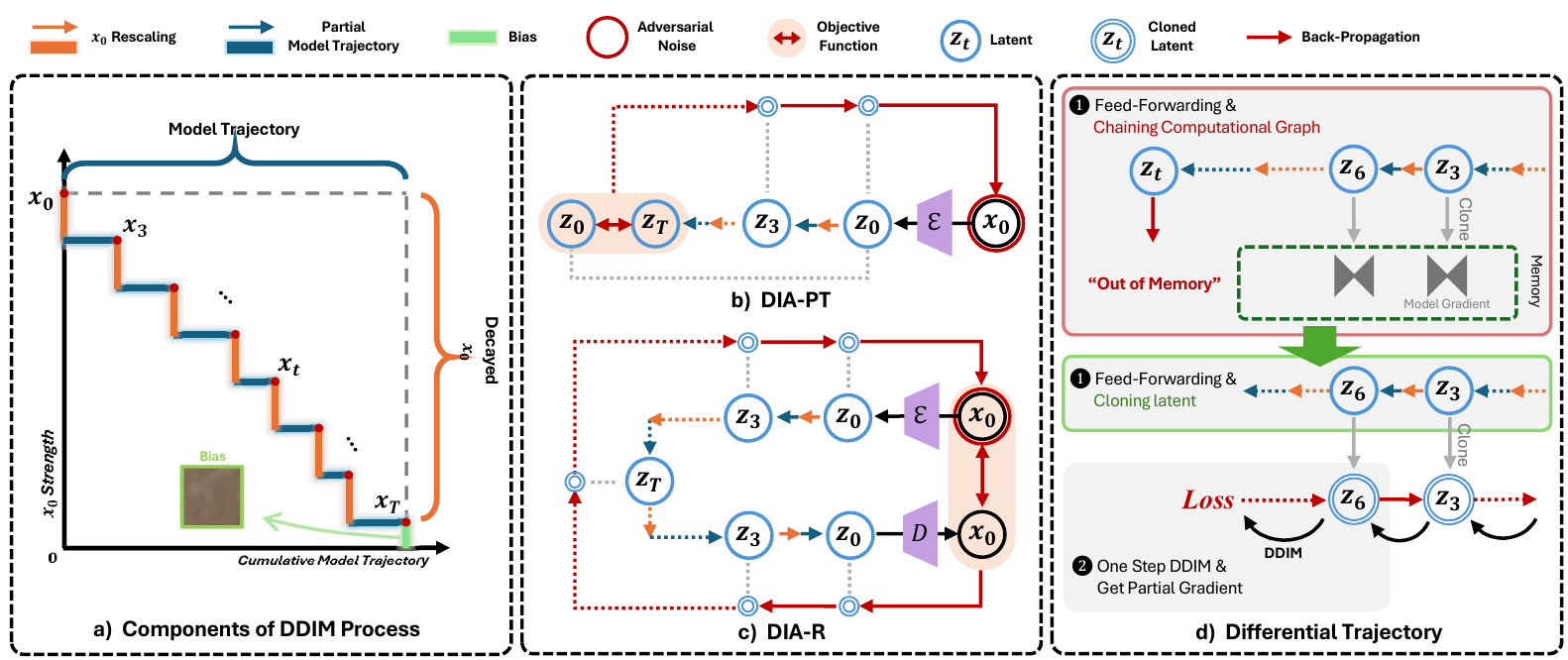}
\caption{\textbf{Overview of the DDIM Inversion Attack Framework (DIA).} In this figure, all items are explained in the context of DDIM with timesteps skipping by 3. a) visualizes the DDIM Process summarized by the strength of $x_0$ and the model trajectory. Each component provides details for explaining the DIA. In b) and c), optimization of $\delta_{\text{DIA-PT}}$ and $\delta_{\text{DIA-R}}$ is shown, which are adversarial noises that interfere with obtaining $z_T$ and $x_0$ using the Differential Trajectory. Finally, d) illustrates the Differential Trajectory to be used in the DDIM process attack. Note that chaining computational graphs to compute the loss results in excessive memory consumption. Therefore, obtaining partial gradients is necessary through step-by-step DDIM inference using cloned latents.
 }
\label{fig:method}
\vspace{-8pt}
\end{figure*}  

\section{Method}
\subsection{Preliminaries}
\subsubsection{Formulation of DDIM Inversion for Latent Interpretation}

To enhance the readability of our paper, we will share the same notation as DDPM~\cite{ho2020denoising}. Let $x_0$ be the clean image, $x_T$ be a space of $\mathcal{N}(0,1)$ when $T=1000$, and $\{ \beta_t \}^{T=1000}_{t=1}$ be a pre-defined noise schedule with $\alpha_t := 1-\beta_t$ and $\bar{\alpha}_t$ := $\prod_{\tau=1}^t \alpha_\tau$. The diffusion forward process can be expressed as:
$$x_t = \sqrt{\bar{\alpha}_t} x_0 + \sqrt{1-\bar{\alpha}_t} \epsilon, \text{ where }\epsilon \sim \mathcal{N}(0, I).$$
With the assumption of linearization between $\epsilon(x_t,t)\approx\epsilon(x_{t-1},t)$, the DDIM inversion process can be shown to be:

\begin{align}\label{eq:ddim_inv}
x_{t+1}&=\sqrt{\alpha_{t+1}}x_t+\underbrace{\sqrt{\bar{\alpha}_{t+1}}(\lambda(t))\epsilon_{\theta}(x_t,t+1)}_{\text{noising part }\Delta_t}.
\end{align}




\subsubsection{Adversarial Attacks}
An adversarial attack is the optimization of the inputs to steer the model's behavior. One widely used attack is the Projected Gradient Descent (PGD) attack~\cite{madry2017towards}, which iteratively adjusts the input data to maximize model loss while ensuring the perturbed data stays within an epsilon-ball around the original point, given by the following formula:
\begin{equation}
x' = \Pi_{x+S}(x + \alpha \cdot \text{sign}(\nabla_x J(\theta, x, y))).
\end{equation}

Here, \( x \) is the original input, \( x' \) is the perturbed data, \( J \) is the model's loss function, \( \theta \) are the model parameters, \( y \) is the true label, \( \alpha \) is the step size, \( \nabla_x J(\theta, x, y) \) is the gradient of the loss, \( \text{sign}(\cdot) \) is the sign function, and \( \Pi_{x+S}(\cdot) \) is the projection operator. Most protection methods uniformly operate under the PGD update. Likewise, our method is similar in this aspect.

\begin{figure*}[h]
\centering
\includegraphics[width=\textwidth]{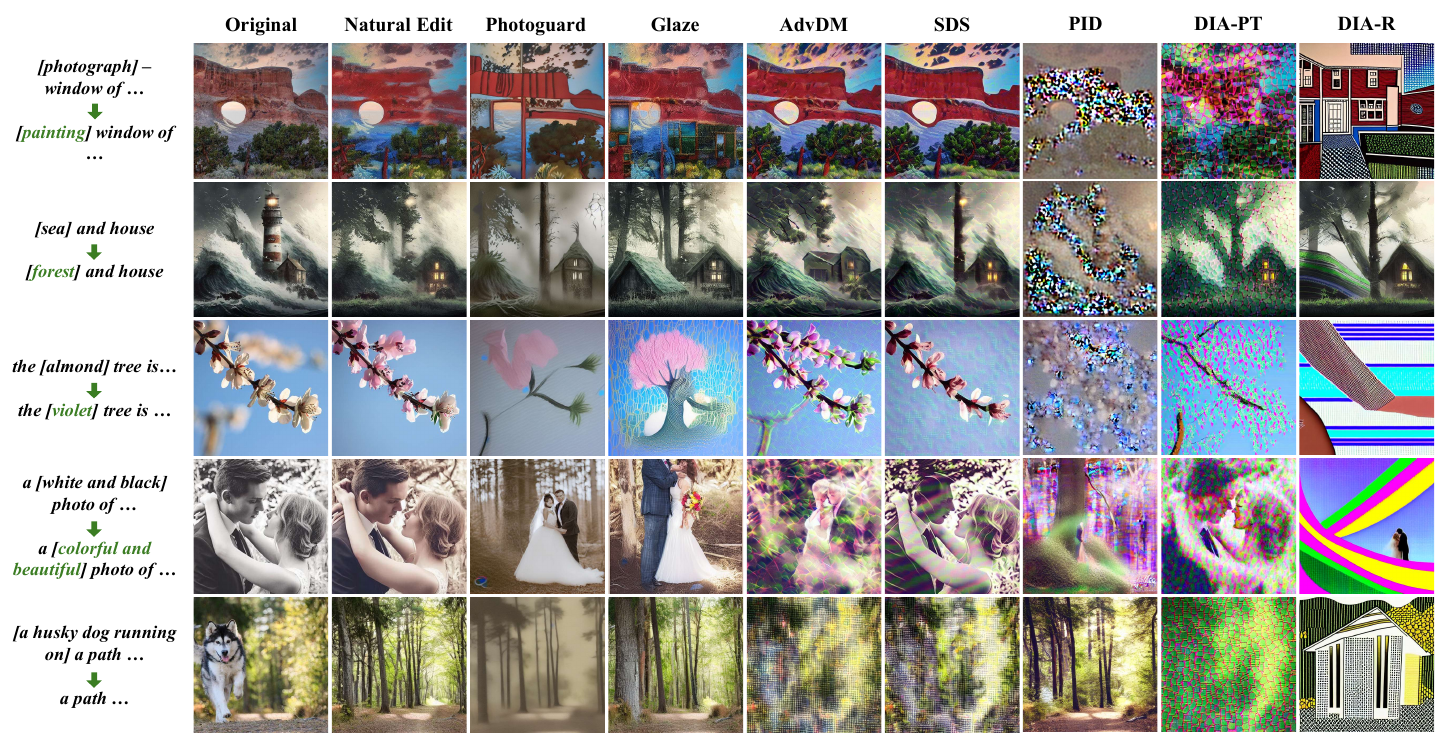}
\caption{\textbf{Quality comparison of images generated by DDIM-to-DDIM across different immunization methods.} The words in green indicate the parts to be edited from the original image. Each method has a different immunization performance compared to Natural Edit. Our method, DIA-PT and DIA-R, demonstrate robust image protection performance on various images.}
\label{fig:qualitative}
\vspace{-8pt}
\end{figure*}

\subsection{Disrupting the Process Trajectory: DIA-PT and DIA-R}
AdvDM~\cite{liang2023adversarial} focuses on attacking the partial trajectory of the diffusion process. However, disrupting this is akin to disrupting the behavior on the noised image, and does not directly target the diffusion chain by design. Our goal is to devise an inversion-oriented objective that is supplementary to existing loss functions.

Methods using DDIM inversion typically strive to obtain a faithful encoding $x_T$, which serves as a departing point for denoising in various editing methods. To interfere with the retrieval of $x_T$ of high fidelity, we first examine the $x_T$ obtained from DDIM inversion, which can be expressed as follows using Eq.~\ref{eq:ddim_inv}:
\begin{align}\label{eq:x_T_mt}
x_{T}&=\underbrace{\sqrt{\bar{\alpha}_{T}}x_0}_{\text{bias}}+\underbrace{\sum^{T}_{i=0}\frac{\sqrt{\bar{\alpha}_{T}}}{\sqrt{\bar{\alpha}_{i+1}}}\Delta_i}_{\text{MT}}
\end{align}

From Eq.~\ref{eq:x_T_mt}, we highlight that $x_T$ is composed of the decayed $x_0$ and the model trajectory (MT). Here, we can obtain the inversion process trajectory built from $x_0$ through a simple  substitution as follows:
\begin{align}\label{eq:x_T_pt}
x_{T}&=x_0+\underbrace{(\sqrt{\bar{\alpha}_{T}}-1)x_0+\sum^{t}_{i=0}\frac{\sqrt{\bar{\alpha}_{T}}}{\sqrt{\bar{\alpha}_{i+1}}}\Delta_i}_{\text{PT}}.
\end{align}

Seamlessly, the complement to $x_0$ equates to Process Trajectory (PT), which encompasses both the decayed $x_0$ and the accumulated model trajectory. This formulation bridges to DIA-PT, by isolating PT:

\begin{equation} \label{eq:DIA_PT}
\delta_{\text{DIA-PT}} = \argmax_{||\delta|| \leq \epsilon} \|\hat{x}_{0:T}(x_0+\delta) - \mathcal{E}(x_0+\delta)\|^2_2 ~,
\end{equation}
 where $\hat{x}_{i:j}$ represents the process of inversion, specifically the transition from timestep $i$ to $j$ during the backward denoising process.
Through Eq.~\ref{eq:DIA_PT}, we can attack the inversion process trajectory to $x_T$. Intuitively, we maximize accumulated changes in the reverse diffusion process to corrupt the inverted latent code used for editing. Attacking the PT provides a more comprehensive attack considering both original image information and model prediction interactions.

Extending from the DIA-PT, we can sample $x_0$ from $x_T$ in a differentiable manner. While we have obtained a differentially chained sample $x_T$, the immediate application of the diffusion loss is not possible, due to the absence of an $\epsilon$ to be predicted in standard diffusion loss. Thankfully, we do have access to the original sample $x_0$, which allows us to reparameterize and predict the clean image from any intermediary $x_t$:
\begin{equation}
\begin{aligned} \label{DIA_R}
\delta_{\text{DIA-R}}=\argmax_{||\delta|| \leq \epsilon}\|\Tilde{x}_{T:0}(\hat{x}_{0:T}(&x_0+\delta)) -(x_0+\delta)\|^2_2
\end{aligned}
\end{equation}
where  $\tilde{x}_{i:j}$ refers to the reconstruction process, which involves generating the data at timestep $j$ from timestep $i$ by following the reverse diffusion steps.

In contrast to AdvDM which relies on a stochastic forward process to yield loss, we sample up to $x_T$ and back to $x_0$ with a deterministic forward and reverse diffusion process, respectively. Intuitively, maximizing Eq.~\ref{DIA_R} implies a digression of the reconstructed image away from the source image.

\subsection{Differentiable Diffusion Trajectory}\label{flowgrad}
Commonly, tracking gradients during the sampling of the trajectory in equations Eq.~\ref{eq:ddim_sampling} or~\ref{eq:ddim_inversion} are impractical due to the significant memory requirements. Fortunately, FlowGrad~\cite{liu2023flowgrad} leverages the decomposition of backpropagation by computing the timestep-wise vector-Jacobian product instead of the entire Jacobian, effectively reducing memory usage as follows:
\begin{equation}
    \nabla_{h_t}\mathcal{J}=
    \begin{cases}
        \frac{\partial\mathcal{L}}{\partial h_{t}}, & t = T \\
        \nabla_{h_{t+1}}\mathcal{J} \cdot J_{\text{VAE}}(h_t), & t = 0 \\
        \nabla_{h_{t+1}}\mathcal{J} \cdot J_{\text{DDIM}}(h_t), & \text{otherwise}
    \end{cases}
\end{equation}
where DDIM is a single step with Eq.~\ref{eq:ddim_sampling} and $J_{\text{DDIM}}(h_t)$ is the Jacobian of DDIM at point $h_t$.  $h_t \in \{h_i\}^{T}_{i=0}$ is $t$-th intermediate point of diffusion trajectory. $T$ is the full trajectory length. And, $J_{\text{VAE}}$ is Jacobian of VAE model and $\nabla_{h_{t+1}}\mathcal{J}$ is accumulated gradient of loss against $h_{t+1}$. 
Effectively, the gradient computation can be decomposed into a sequence of decoupled intermediate gradient calculations, where each of them can be yielded without storing the activations of the full trajectory.

%% file: sec/5_Experiment.tex
\begin{table*}[ht]
\fontsize{9}{11}\selectfont
\centering
\setlength{\tabcolsep}{0.7mm}{
\begin{tabular}{c|cccc|cc|cc|c}
\toprule
\textbf{Inversion}   & \multicolumn{4}{c|}{\textbf{DDIM Inversion}}               & \multicolumn{2}{c|}{\textbf{Null-Text Inversion}} & \multicolumn{2}{c|}{\textbf{Negative-Prompt Inversion}} & \textbf{Direct Inversion}  \\
\toprule
\textbf{Edit}        & \textbf{DDIM}    & \textbf{MasaCtrl} & \textbf{PnP}     & \textbf{P2P}     & \textbf{P2P}      & \textbf{Proximal-Guidance}  & \textbf{P2P}      & \textbf{Proximal-Guidance} & \textbf{P2P}     \\
\toprule
Natural Edit
 & 25.7100 & 24.9504 & 26.1413 & 25.9123 
 & 25.5750 & 24.8495 & 25.4566 & 25.2090 
 & 25.8333 \\
\midrule
PhotoGuard
 & 24.6400 & 22.8856 & 24.7364 & \textbf{25.9267}
 & 24.0286 & 22.8213 & 21.6895 & 21.3095
 & \underline{26.0429} \\

Glaze
 & 25.5147 & 23.8529 & 26.0200 & \underline{25.9394}
 & 25.5676 & 24.2446 & 24.0998 & 23.8052
 & 26.6814 \\

AdvDM
 & 24.5179 & 22.3192 & 23.2544 & 26.1522
 & 23.7018 & 21.4290 & 18.9884 & 18.7983
 & 26.2887 \\
 

SDS
 & 24.2051 & 23.1265 & 23.4413 & 25.9414
 & 24.0519 & 21.7499 & 19.8636 & 19.7851
 & \textbf{25.7531} \\

PID
 & \textbf{21.2091} & 23.8213 & 25.6779 & 25.9553
 & 24.8942 & 23.6791 & 23.2155 & 22.9292
 & 26.9447 \\

\rowcolor[gray]{0.9} DIA-PT (ours)
 & \underline{23.4614} & \textbf{18.3076} & \underline{20.7749} & 26.0381
 & \underline{23.1999} & \underline{20.0267} & \underline{17.4938} & \underline{17.3992}
 & 26.0563 \\
 
\rowcolor[gray]{0.9} DIA-R (ours)
 & 23.4626 & \underline{19.3155} & \textbf{18.4336} & 26.0173
 & \textbf{22.3095} & \textbf{18.7471} & \textbf{15.0552} & \textbf{14.8728}
 & 26.3062 \\
\bottomrule
\end{tabular}
\caption{\textbf{CLIP similarity between the edited image and the prompt:} Under a combination of different image inputs (original or immunized) and an inversion-editing method pairing, we show the CLIP similarity for images in the PIE-Bench dataset. Lower CLIP similarity indicates better immunization.}
\label{tab:metrics}
}
\end{table*}





\section{Experiment} 
\subsection{Experiment Details}

\paragraph{Datasets} We evaluate the attack using the PIE benchmark~\cite{ju2023direct}. PIE bench is a benchmark designed to assess image editing performance, consisting of 700 images divided into 9 sub-tasks (e.g. changing object, deleting object and changing style), enabling a total of 6,300 evaluations. The dataset includes images along with source prompts, target prompts, editing instructions, and editing masks. Here, the editing mask bounds the portion of the source image where editing should take place, with the inside of the mask as the foreground and the outside of the mask as the background. While this dataset was originally designed to evaluate the quality of various inversion and image editing methods, the quantification of disruption methods' success exhibits an inverse relationship with editing quality metrics within the PIE benchmark. Hence, we focus on identifying the worst-performing attack method under the PIE bench.


\paragraph{Evaluation Metrics} 
Disruption assessment, like benchmarking image edits, evaluates identity, structure preservation, and visual realization of the edit prompt.
In PIE-bench, the integrated metrics allow us to precisely assess background preservation (with \textbf{PSNR}, \textbf{LPIPS}, \textbf{MSE}, and \textbf{SSIM})~\cite{zhang2018unreasonable, 1284395} and prompt-image consistency (with \textbf{CLIP Similarity})~\cite{radford2021learning, wu2021godiva}.
However, CLIP similarity primarily captures semantic coherence, potentially yielding high scores even for severely distorted images. We therefore employ auxiliary metrics (PSNR, LPIPS, SSIM) to complement evaluation. While CLIPScore is biased toward providing ``credit for being roughly right,'' a markedly low score provides strong evidence of a genuine misalignment.

Additionally, PIE-Bench incorporates isolation of the edited portion through masking, which we utilize when evaluating background preservation. In the case of CLIP similarity, we compare the cosine similarity between the unmasked image and the edit text embedding to accurately reflect the context of the image. 

\paragraph{Attack Baselines and Setup} 
We compare our approach with Photoguard~\cite{salman2023raising}, Glaze~\cite{shan2023glaze}, AdvDM~\cite{liang2023adversarial}, SDS~\cite{xue2023toward} and PID~\cite{li2024pid} as a baseline to DIA-R and DIA-PT, which attack the trajectory of our proposed methodology. Glaze, Photoguard and PID are methodologies that attack VAEs, while AdvDM and SDS are designed to attack diffusion models without considering trajectories.

The inversion methods considered are DDIM inversion, Direct Inversion, Negative-Prompt Inversion, and null-text Inversion~\cite{dhariwal2021diffusion, miyake2023negative, mokady2023null}. For the editing methods, we test on the plain DDIM reconstruction, Prompt-to-Prompt (P2P), Proximal-Guidance, and MasaCtrl~\cite{hertz2022prompt, han2023improving, cao2023masactrl}.
Since image editing requires combining an inversion and editing method, we select 9 representative pairings (e.g. DDIM-to-DDIM, Negative-to-P2P) that capture their distinct behaviors, as shown in Table~\ref{tab:metrics}. All edits are performed on Stable Diffusion v1.4~\cite{rombach2022high} using default benchmark settings.
The settings for the immunization methods used in the experiment are as follows: All methods use a PGD~\cite{madry2017towards} perturbation epsilon of 0.05. The iterations for Photoguard and AdvDM are set to 60, while DIA-PT and DIA-R use 20 iterations to train the adversarial noise. Additionally, the inversion and reconstruction process trajectories used in DIA-PT and DIA-R each consist of 10 DDIM steps.

\subsection{Qualitative Results}
The task of real-image editing is inherently subjective. Moreover, we have presented different method variants, where each of them is uniquely motivated. Therefore, we present each disruption method's results on a wide range of edits including style changing, object changing, and object deletion, shown in Fig~\ref{fig:qualitative}.

In concordance with previous works~\cite{wang2023simac, liang2023adversarial}, Photoguard, Glaze and PID show varying performance results. These results rather highlight the repairing capability of current large-scale text-to-image diffusion models; even when provided with a corrupted initial $x_0$, the models can effectively steer the adversarial diffusion path back to the original path. 

Indifferently, AdvDM and SDS display inconsistent success across different images. We hypothesize that this is due to their multi-timestep constrained optimization, which attacks randomly sampled timestep while only considering partial diffusion trajectory. In contrast, our trajectory-based methods DIA-PT and DIA-R manifest consistent success disruption. Notably, DIA-R demonstrates superior performance as it accumulates residual error throughout the entire learned diffusion process.

\subsection{Quantitative Results}
In this section, we provide some insights and analysis of the quantitative results from PIE-Bench. Here, our two main focuses are background preservation and quantification of the visual substance of our edit prompt in the image. The former ensures that the user has control over the editing region, while the latter ensures that this region indeed receives substantial text-aligned edits. Additional examples and comparisons are provided in Suppl.

\input{sec/preservation_table}

\subsubsection{Comparing Methods through Editing Methods}
We evaluate the efficacy of each immunization method by assessing the number of Inversion-Edit combinations it can impede. Table~\ref{tab:metrics} presents the CLIP Similarity scores between the edited images and their corresponding edit prompts.

Photoguard, Glaze, and other baselines, which do not utilize the DDIM trajectory, show sub-optimal attack performance. Similarly, PID is only effective under DDIM-to-DDIM. In contrast, DIA-PT and DIA-R, which target the chained trajectory, achieve stronger and more transferable attacks across various inversion-editing pairs. This highlights the importance of targeting the inversion trajectory for effective DDIM inversion. 

In particular, for DDIM-to-P2P and Direct-to-P2P, CLIP Similarity is always lower for natural edits than for immunized images. This occurs because these pairings are overly aggressive and fail to preserve original content even in natural images. Additional qualitative results are provided in Suppl.~\ref{sec:ddim_p2p}. 

\subsubsection{Comparing Content Preservation across Immunization Methods}
 
In Table~\ref{tab:preservation_metrics}, we present a comprehensive analysis utilizing multiple metrics to evaluate the mean structure and background preservation of edited images, comparing both clean and immunized ones. 
Our results show that DIA-PT and DIA-R demonstrate a dramatic improvement in disrupting performance, compared to baselines. As evidenced in our qualitative results, DIA-PT imprints a uniform synthetic artifact spread across the image. This characteristic leads to outstanding performance in metrics that evaluate perceptual similarity, such as LPIPS and SSIM. However, DIA-R shows strength in metrics that measure pixel-wise differences, such as PSNR and MSE.

\begin{figure}[t]
\centering
\includegraphics[width=\columnwidth]{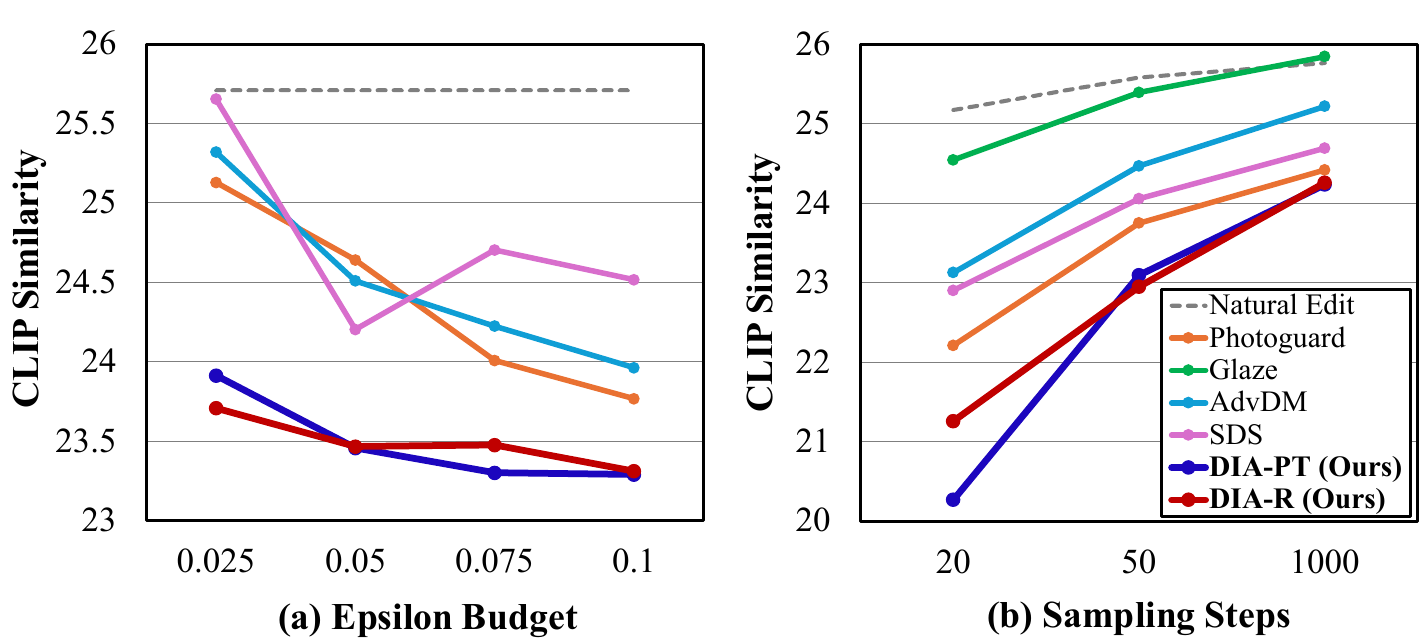}
\caption{\textbf{Comparison of CLIP similarity across different immunization methods through varying epsilon budgets and sampling step levels:} (a) shows CLIP Similarity by Epsilon Budget, and (b) shows CLIP Similarity by Sampling Steps. Lower CLIP similarity indicates better immunization.
}
\label{fig:noise_budget}
\vspace{-8pt}
\end{figure}

\subsection{Evaluating Flexibility on Adversarial Scenarios}


\subsubsection{Comparing Performance Through Noise Budget}
We evaluate the robustness of immunization methods under different noise budget constraints, where the noise budget refers to the maximum norm of adversarial noise added to the image. This noise budget balances image quality and attack efficacy. Our evaluation in Fig.~\ref{fig:noise_budget}~(a) indicates that both DIA-R and DIA-PT consistently outperform baseline methods at all levels of noise budget tested, demonstrating their ability to provide users with high visual quality and protective performance.

\subsubsection{Comparing Performance Through Sampling Steps}
The performance of adversarial attacks is also influenced by the number of sampling steps used during the DDIM-based image editing process. Increasing sampling steps generally weakens attack effectiveness since smaller adversarially perturbed denoising steps are less consequential to future steps, and thus allow room for repair. This parameter is particularly relevant as users often adjust sampling steps based on practical constraints or quality requirements.

To assess robustness against varying sampling settings, we evaluate attack performance using commonly employed sampling steps: shorter (20 steps), standard (50 steps), and extended (1,000 steps) as shown in Fig.~\ref{fig:noise_budget}~(b). Our results demonstrate that attack effectiveness significantly decreases as sampling steps increase. Notably, our proposed methods, DIA-R and DIA-PT, maintain robust attack performance even at the challenging setting of 1,000 sampling steps, highlighting their superior adaptability and resilience across diverse sampling scenarios.

\begin{figure}[t]
\centering
\includegraphics[width=\columnwidth]{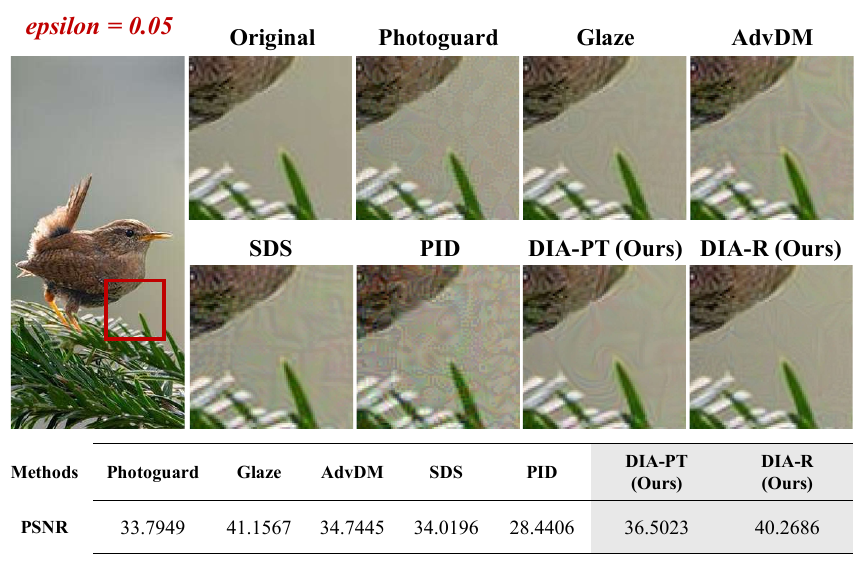}
\caption{\textbf{Comparison of image degradation levels across immunization methods under perturbation epsilon = 0.05.} The top row visualizes adversarial noise, where cleaner indicates better performance. The bottom shows average PSNR values measured between 700 pairs of original and immunized images from PIE-Bench. Higher PSNR indicates better stealthiness.
}
\label{fig:advnoise}
\vspace{-8pt}
\end{figure}

\subsubsection{Comparing Performance Through Purification}
\begin{figure}[t]
\centering
\includegraphics[width=\columnwidth]{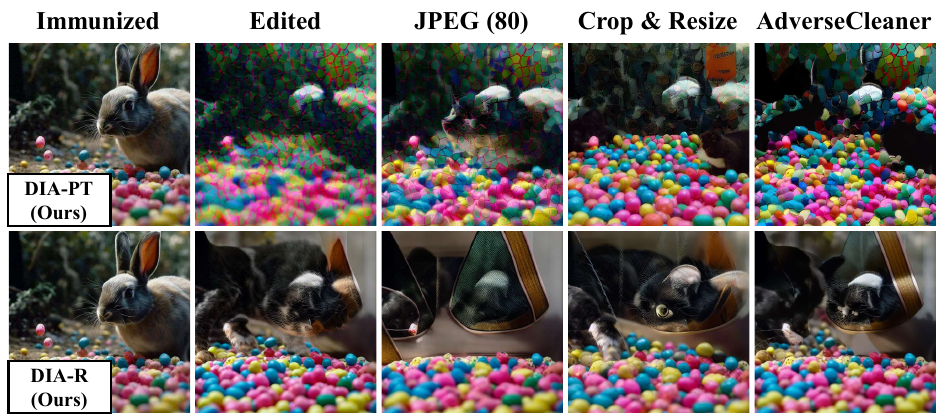}
\caption{\textbf{Visualization of robustness through purification methods.}  The first and second rows show the edited images for DIA-PT and DIA-R, respectively. The second column displays the edited images of immunized images without purification, while the remaining columns show the edited images after applying purification to the immunized images. The editing task is \textit{``a [rabbit \textcolor{OliveGreen}{$\rightarrow$ cat}] is sitting in a pile of colorful eggs."}
}
\label{fig:purification_figure}
\vspace{-8pt}
\end{figure}

Immunization methods are always exposed to purification, whether intended or not. We evaluated robustness through commonly accessible purification techniques such as JPEG Compression, Crop \& Resize, and Adverse Cleaner~\cite{advclean}. As shown in Fig.~\ref{fig:purification_figure}, our method maintains strong performance, with only minor degradation. Extensive comparison is provided in Suppl.~\ref{sec:suppl_noise_purification}.

\subsection{Comparing Perturbed Images across Immunization Methods}
We also emphasize that our methods are distinguished from previous works concerning synthesized immunization noise under the epsilon budget of 0.05. 
As shown in Fig.~\ref{fig:advnoise}, Photoguard exhibits a uniform ``scale" pattern, while AdvDM, SDS, PID and DIA-PT leave low-frequency types of patterns. In contrast, Glaze and DIA-R demonstrate stealthiness at such a level where it is difficult to find distinctive patterns. To demonstrate our observations, we present average PSNR values measured between 700 pairs of original and immunized images from PIE-Bench in the table shown in Fig.~\ref{fig:advnoise}. Notably, DIA-R showed closest PSNR values to Glaze, which also secures visual imperceptibility as an objective. These results highlight DIA-R's strong stealthiness achieved without a dedicated objective.

%% file: sec/preservation_table.tex
\begin{table}[t]
\fontsize{9}{11}\selectfont
\centering
\setlength{\tabcolsep}{1mm}{
\begin{tabular}{c|c|cccc}
\toprule
\textbf{Metrics} & \textbf{Structure} & \multicolumn{4}{c}{\textbf{Background Preservation}} \\ 
\midrule
\textbf{Method}         
 & \textbf{Distance $\uparrow$} 
 & \textbf{PSNR $\downarrow$} 
 & \textbf{LPIPS $\uparrow$} 
 & \textbf{MSE $\uparrow$} 
 & \textbf{SSIM $\downarrow$}   \\
\toprule
Natural Edit 
 & 0.0249 
 & 24.3767 
 & 0.0914 
 & 0.0071 
 & 0.8124 \\
\midrule
PhotoGuard 
 & 0.0773 
 & 19.6509 
 & 0.2617 
 & 0.0148 
 & 0.6584 \\

 Glaze 
 & 0.0440 
 & 21.3841 
 & 0.1927 
 & 0.0111 
 & 0.6958 \\
 
 AdvDM 
 & 0.0940 
 & 19.6309 
 & 0.2838 
 & 0.0167 
 & 0.5933 \\

 SDS 
 & 0.0685 
 & 20.5587 
 & 0.2703 
 & 0.0135 
 & 0.6232 \\

 PID 
 & 0.0630 
 & 20.0265 
 & 0.2878 
 & 0.0151 
 & 0.6211 \\

\rowcolor[gray]{0.9} DIA-PT (ours)
 & \underline{0.1059} 
 & \underline{18.2202} 
 & \textbf{0.3410} 
 & \underline{0.0237} 
 & \textbf{0.5653} \\
\rowcolor[gray]{0.9} DIA-R (ours)
 & \textbf{0.1252} 
 & \textbf{16.3055} 
 & \underline{0.2940} 
 & \textbf{0.0460} 
 & \underline{0.5903} \\
\bottomrule
\end{tabular}
}
\caption{\textbf{Average background and structure preservation metric for 9 editing techniques.} This metric assesses how well the unedited regions are preserved.}
\label{tab:preservation_metrics}
\end{table}

%% file: sec/6_Discussion.tex
\section{Discussion \& Limitation}

DIA is a practical method that protects images shared online from being maliciously edited to spread misinformation or used without permission. By immunizing images using the proposed method before they are shared online, it can be utilized to prevent the spread of misinformation by suppressing malicious image editing techniques while minimizing image degradation.




Although our method shows promising results, we acknowledge the inherent limitation of cross-model transferability and vulnerability to noise purification approaches that are shared by existing image immunization methods. We provide experimental validation in the Suppl.~\ref{sec:suppl_black_box},  ~\ref{sec:suppl_noise_purification}.

Also, further experimental details and numerical analyses are available in the Supplementary Materials.


  %

%% file: sec/7_Conclusion.tex
\section{Conclusion}

We propose DDIM Inversion Attack (DIA) to disrupt DDIM Inversion-based editing methods. First, we provide DDIM trajectory-based attack variants DIA-PT and DIA-R by exploiting potential vulnerabilities in DDIM Inversion. Namely, we highlight that the integration of the differentiable DDIM trajectory into the objectives enhances their disruption capability. Through extensive experiments on a collection of editing methods across, we demonstrate that our algorithms are both efficient and effective. In summary, our work contributes to practically preventing further privacy threats as well as malicious use by providing an immunization technique that captures the underlying mechanism of SoTA editing methods.

\section*{Acknowledgments}
This work was partly supported by Institute for Information \& communication Technology Planning \& evaluation (IITP) grants funded by the Korean government MSIT:
(RS-2022-II221199, RS-2022-II220688, RS-2019-II190421, RS-2023-00230337, RS-2024-00356293, RS-2024-00437849, RS-2021-II212068,  RS-2025-02304983, and  RS-2025-02263841).

%% file: sec/99_Appendix.tex
\onecolumn
\clearpage
\setcounter{page}{1}
\setcounter{section}{0}
\setcounter{figure}{0}
\setcounter{equation}{0}

{\centering
\large\textbf{\thetitle} \\ 
\vspace{0.5em}Supplementary Material \\
\vspace{1.0em}
}


\section{Disrupting Deterministic Inversion with Differentiable Trajectory: DIA-MT}
We aimed to attack the inversion process by maximizing the process trajectory (PT), which is derived from the difference between the initial point $x_0$ and the final point $x_T$. However, it is also valid to target only the model's predicted trajectory (MT). This MT is derived as shown in Equation 7, and can be understood as exclusively capturing the contribution of the model's predictions to the inversion process. 
We propose an attack on this MT, called DIA-MT, which maximizes the residual image signal, defined as ($x_T$ $-$ decayed $x_0$), away from an isotropic Gaussian. DIA-MT is formulated as follows:
\begin{equation} \label{eq:DIA_MT} 
\delta_{\text{DIA-MT}} = \argmax_{||\delta|| \leq \epsilon} \|\hat{x}_{T}(x_0+\delta)-\sqrt{\bar{\alpha}_{T}}(\mathcal{E}(x_0+\delta))\|^2_2
\end{equation}
Same as DIA-PT, $x_0+\delta$ is detached from the computational graph used to calculate the gradient.
Additionally, as an ablation study for DIA-MT, we compared its background preservation and prompt-image consistency with those of DIA-PT in Table~\ref{tab:mt_background}. Here, the "Natural Edit" represents the natural outcome of an image editing process without any disruption, and it is used as a reference point in our experiments.
\begin{table*}[ht]
\fontsize{9}{11}\selectfont
\centering
\setlength{\tabcolsep}{0.9mm}{
\begin{tabular}{c|cccc|cc|cc|c}
\toprule
\textbf{Inversion}   & \multicolumn{4}{c|}{\textbf{DDIM Inversion}}               & \multicolumn{2}{c|}{\textbf{Null-Text Inversion}} & \multicolumn{2}{c|}{\textbf{Negative-Prompt Inversion}} & \textbf{Direct Inversion}  \\
\toprule
\textbf{Edit}        & \textbf{DDIM}    & \textbf{MasaCtrl} & \textbf{PnP}     & \textbf{P2P}     & \textbf{P2P}      & \textbf{Proximal-Guidance}  & \textbf{P2P}      & \textbf{Proximal-Guidance} & \textbf{P2P}     \\
\toprule
Natural Edit & 25.7100 & 24.9504  & 26.1414 & 25.9123 & 25.5750  & 24.8495            & 25.4566  & 25.2090           & 25.8333 \\
\midrule
DIA-PT        & \textbf{23.4614}& \textbf{18.3076}& \textbf{20.7749} & 26.0381 & \textbf{23.1999}& \textbf{20.0266}            & \textbf{17.4938}  & \textbf{17.3992}        & 26.0563\\
DIA-MT      & 23.7177 & 21.8592  & 23.4419 & \textbf{25.7381} & 24.4444  & 22.6471   & 21.4318  & 21.2247           & \textbf{25.4861} \\

\bottomrule
\end{tabular}
\caption{CLIP similarity between the edited image and the prompt: Under a combination of different image inputs (clean or disrupted) and an inversion-editing method pairing, we show the CLIP similarity for images in the PIE-Bench dataset. Lower CLIP similarity indicates better immunization.
}\label{tab:mt_clip}
}
\end{table*}
\begin{table}[ht]
\fontsize{9}{11}\selectfont
\centering
\setlength{\tabcolsep}{1mm}{
\begin{tabular}{c|c|cccc}
\toprule
\textbf{Metrics}        & \textbf{Structure}          & \multicolumn{4}{c}{\textbf{Background Preservation}} \\ 
\toprule
\textbf{Method}         & \textbf{Distance $\uparrow$} & \textbf{PSNR $\downarrow$} & \textbf{LPIPS $\uparrow$} & \textbf{MSE $\uparrow$} & \textbf{SSIM $\downarrow$}   \\
\toprule
Natural Edit & 0.0249             & 24.3767 & 0.0914 & 0.0071 & 0.8124 \\
\midrule
DIA-PT         & \textbf{0.1059}    & \textbf{18.2202} & \textbf{0.3410} & \textbf{0.0237} & \textbf{0.5653} \\
DIA-MT         & 0.0514    & 22.0443 & 0.2447 & 0.0107 & 0.6856 \\
\bottomrule
\end{tabular}
\caption{Average background and structure preservation metric for 9 editing techniques. This metric assesses how well the unedited regions are preserved.}\label{tab:mt_background}
}

\end{table}
\FloatBarrier 

According to Table~\ref{tab:mt_clip} and Table~\ref{tab:mt_background}, DIA-MT showed that attacking model trajectories are indeed effective, supporting our main argument that trajectories should be taken into account during attacks. However, it is observed that the performance of DIA-MT, which excludes the scaling of $x_0$ during the inversion process, is slightly weaker compared to DIA-PT, which includes it. This suggests that considering the scaling of $x_0$ leads to a more effective attack.

\newpage
\section{More experimental results}

\subsection{Step Generalizability}
The number of steps used in DDIM varies according to the user's preference and budget. However, in DIA-PT and DIA-R, we execute the attack with trajectories sampled with 10 DDIM steps during the inversion and reconstruction process. Therefore, it is important to verify that our method works in editing environments using different timestep spacings. In Table~\ref{tab:step_gen}, we compared the performance by setting DDIM steps to 20, 50, 200, and 1000 in an editing environment DDIM-to-DDIM for 140 randomly selected images from PIE-Bench~\cite{ju2023direct}. 

\begin{table*}[ht]
\fontsize{9}{11}\selectfont
\centering

\setlength{\tabcolsep}{4.2mm}{
\begin{tabular}{c|c|cccccc}
\toprule

                         \textbf{Inference Step}&        \textbf{Method}& \textbf{CLIP}$\downarrow$ & \textbf{Distance $\uparrow$} & \textbf{PSNR $\downarrow$} & \textbf{LPIPS $\uparrow$} & \textbf{MSE $\uparrow$} & \textbf{SSIM $\downarrow$}  \\ 
\toprule
\multirow{7}{*}{20} & Natural Edit  &  25.1773&  0.02102&  25.6405&  0.07350&  0.00460&  0.83539\\
                         & Photoguard  &  22.2119&  0.08438&  20.0121&  0.26683&  0.01327&  0.65680\\
                         & Glaze  &  24.5483&  0.04575&  21.8093&  0.20853&  0.01063&  0.67463\\
                         & AdvDM  &  23.1290&  0.08768&  20.4630&  0.27519&  0.01470&  0.60280\\
                         & SDS  &  22.9042&  0.06344&  21.0721&  0.26125&  0.01203&  0.62612\\
                         \rowcolor[gray]{0.9}     \cellcolor{white} & DIA-PT (ours) &  \textbf{20.2686}&  \textbf{0.13860}&  \textbf{17.1762}&  \textbf{0.38523}&  0.02803&  \textbf{0.51990}\\
                         \rowcolor[gray]{0.9}     \cellcolor{white} & DIA-R (ours)  &  21.2578&  0.11058&  17.5010&  0.28155&  \textbf{0.04004}&  0.61981\\
 \cmidrule{1-8}
\multirow{7}{*}{50} & Natural Edit  &  25.5855&  0.02488&  24.7736&  0.08818&  0.00566&  0.82333\\
                         & Photoguard  &  23.7524&  0.08060&  19.4990&  0.24705&  0.01432&  0.68406\\
                         & Glaze  &  25.3976&  0.04158&  22.1291&  0.18450&  0.00979&  0.71290\\
                         & AdvDM  &  24.4719&  0.07046&  21.0579&  0.23121&  0.01242&  0.66378\\
                         & SDS  &  24.0583&  0.06431&  21.1631&  0.22512&  0.01174&  0.66120\\
                         \rowcolor[gray]{0.9}     \cellcolor{white} & DIA-PT (ours) &  23.0969&  \textbf{0.10368}&  \textbf{19.0090}&  \textbf{0.31325}&  0.02019&  \textbf{0.59189}\\
                         \rowcolor[gray]{0.9}     \cellcolor{white} & DIA-R (ours)  &  \textbf{22.9501}&  0.08432&  19.2288&  0.22481&  \textbf{0.02689}&  0.68966\\
\cmidrule{1-8}
\multirow{7}{*}{1000} & Natural Edit  &  25.7686&  0.02837&  23.9500&  0.10169&  0.00668&  0.81311\\
                         & Photoguard  &  24.4224&  \textbf{0.07838}&  \textbf{19.5179}&  0.24314&  0.01427&  0.70117\\
                         & Glaze  &  25.8500&  0.03955&  21.9824&  0.16913&  0.00955&  0.73640\\
                         & AdvDM  &  25.2232&  0.06341&  21.2205&  0.20878&  0.01198&  0.69459\\
                         & SDS  &  24.6955&  0.06262&  21.2171&  0.20686&  0.01153&  0.69147\\
                         \rowcolor[gray]{0.9}     \cellcolor{white} & DIA-PT (ours) &  \textbf{24.2379}&  0.07757&  20.1859&  \textbf{0.25300}&  0.01587&  \textbf{0.65586}\\
                         \rowcolor[gray]{0.9}     \cellcolor{white} & DIA-R (ours)  &  24.2615&  0.06724&  19.8701&  0.19693&  \textbf{0.01990}&  0.72183\\
\bottomrule
\end{tabular}
\caption{Attack Performance Across Different Editing Steps. This table shows the performance of various attack methods using 20, 50, and 1000 DDIM steps for inversion and reconstruction on 140 images from the PIE-Bench dataset. Key metrics include CLIP Similarity (CLIP), Distance, PSNR, LPIPS, MSE, and SSIM.}
\label{tab:step_gen}
}
\end{table*}
\FloatBarrier 
A notable observation from these results is that the attack performance decreases as the number of steps increases, which is evident across all metrics. Our experiments include assessments with 1000 steps, the maximum step size typically used in the diffusion process, where we observe the poorest attack performance. However, it is crucial to note that the performance remains consistently lower than that of natural edits performed without any attack. This demonstrates the efficacy of our method across all step sizes and supports the stability of our approach.

\newpage
\subsection{Comparing Performance Through Noise Purification}\label{sec:suppl_noise_purification}
In this section, we compare the robustness of different methods against cleaning approaches known as `purification' for adversarial noise. We provide performance measurements after applying JPEG Compression, Crop \& Resize, and AdverseCleaner to 700 immunized images across all methods. Details for each purification method are as follows:

\begin{itemize}
    \item 
        JPEG Compression: The simplest and fastest image compression algorithm for purifying adversarial noise. Compression quality can be selected between 0 and 100, where lower values cause more image degradation. We provide results with quality values of 70, 80, and 90.
        
    \item 
        Crop \& Resize: A naturally occurring and effective purification technique. We cropped 10\% of each image and then resized it to match the model's input requirements.

    \item 
        Adverse Cleaner~\cite{advclean}: An algorithmic approach capable of purifying high-frequency noise patterns.

    \item
        Gaussian Noising: The purification method that adds random Gaussian noise on immunized images. We provide results with $\sigma$=0.1.
        
    \item 
        Noisy Upscaling~\cite{honig2024adversarial}: A two-stage purification method proposed by \citet{shan2023glaze}, which applies Gaussian Noising ($\sigma$=0.1) followed by Stable Diffusion Upscaler~\cite{rombach2022high}.

\end{itemize}

As shown in Table~\ref{tab:purification}, all baselines demonstrate robustness to purification when compared to Natural Edit. Notably, our method maintains superior performance while remaining robust to most purification methods. In some experiments, SDS shows sub-optimal performance, which appears to be due to its low-frequency pattern and higher degradation scale.

\newpage

\begin{table*}[ht]
\fontsize{9}{11}\selectfont
\centering

\setlength{\tabcolsep}{3.5mm}{
\begin{tabular}{c|c|cccccc}
\toprule

                         \textbf{Purification Method}&        \textbf{Attack Method}& \textbf{CLIP}$\downarrow$ & \textbf{Distance $\uparrow$} & \textbf{PSNR $\downarrow$} & \textbf{LPIPS $\uparrow$} & \textbf{MSE $\uparrow$} & \textbf{SSIM $\downarrow$}  \\ 
\toprule
\multirow{1}{*}{-} & Natural Edit  &  25.7100&  0.02613&  23.8400&  0.09933&  0.00639&  0.80723\\

 \cmidrule{1-8}
\multirow{6}{*}{JPEG Compression (90)} & Photoguard  &  25.6936&  0.05174&  21.4936&  0.22142&  0.00962&  0.70957\\
                         & Glaze  &  25.8862&  0.03472&  22.4310&  0.15541&  0.00829&  0.74462\\
                         & AdvDM  &  24.5583&  0.07191&  21.0165&  0.21653&  0.01233&  0.67113\\
                         & SDS  &  24.1685&  0.06742&  20.8744&  0.21799&  0.01181&  0.67212\\
                        \rowcolor[gray]{0.9}     \cellcolor{white}  & DIA-PT (Ours) &  24.2789&  0.07655&  20.0105&  \textbf{0.27472}&  0.01610&  \textbf{0.63854}\\
                         \rowcolor[gray]{0.9}     \cellcolor{white} & DIA-R (Ours)  &  \textbf{23.7255}&  \textbf{0.08374}&  \textbf{19.2542}&  0.21633&  \textbf{0.02637}&  0.67706\\
\cmidrule{1-8}
\multirow{6}{*}{JPEG Compression (80)} & Photoguard  &  26.0247&  0.04463&  22.0655&  0.18531&  0.00880&  0.73312\\
                         & Glaze  &  26.0196&  0.03095&  23.0004&  0.13806&  0.00741&  0.76862\\
                         & AdvDM  &  24.4738&  0.07044&  21.1526&  0.21349&  0.01209&  \textbf{0.67737}\\
                         & SDS  &  24.1725&  0.06773&  21.0680&  0.21292&  0.01159&  0.67927\\
                         \rowcolor[gray]{0.9}     \cellcolor{white} & DIA-PT (Ours) &  24.8818&  0.05645&  20.9928&  \textbf{0.23660}&  0.01259&  0.68420\\
                         \rowcolor[gray]{0.9}     \cellcolor{white} & DIA-R (Ours)  &  \textbf{24.2000}&  \textbf{0.07318}&  \textbf{19.9076}&  0.20011&  \textbf{0.02179}&  0.69623\\
\cmidrule{1-8}
\multirow{6}{*}{JPEG Compression (70)} & Photoguard  &  26.0953&  0.04212&  22.3060&  0.16901&  0.00836&  0.74815\\
                         & Glaze  &  26.0306&  0.03010&  23.1220&  0.13043&  0.00712&  0.77931\\
                         & AdvDM  &  24.7252&  \textbf{0.06771}&  21.3877&  0.20781&  0.01172&  0.68590\\
                         & SDS  &  \textbf{24.2350}&  0.06743&  21.1680&  \textbf{0.21060}&  0.01171&  \textbf{0.68238}\\
                         \rowcolor[gray]{0.9}     \cellcolor{white} & DIA-PT (Ours) &  25.5055&  0.04608&  21.6928&  0.20653&  0.01036&  0.71623\\
                         \rowcolor[gray]{0.9}     \cellcolor{white} & DIA-R (Ours)  &  25.0112&  0.06244&  \textbf{20.6276}&  0.18408&  \textbf{0.01673}&  0.71640\\
\cmidrule{1-8}
                         \multirow{6}{*}{Crop \& Resize} & Photoguard  &  25.7733&  0.08424&  17.3266&  0.25859&  0.02362&  0.61375\\
                         & Glaze  &  25.8354&  0.06285&  17.3832&  0.23109&  0.02411&  0.61677\\
                         & AdvDM  &  25.1026&  0.07810&  17.1060&  0.27175&  0.02563&  \textbf{0.57034}\\
                         & SDS  &  \textbf{24.5399}&  0.07795&  16.9971&  0.26720&  0.02630&  0.57432\\
                         \rowcolor[gray]{0.9}     \cellcolor{white} & DIA-PT (Ours) &  24.8340&  0.07498&  16.9972&  \textbf{0.29035}&  0.02618&  0.57572\\
                         \rowcolor[gray]{0.9}     \cellcolor{white} & DIA-R (Ours)  &  24.8310&  \textbf{0.08518}&  \textbf{16.5469}&  0.25361&  \textbf{0.03056}&  0.59598\\
\cmidrule{1-8}
\multirow{6}{*}{Adverse Cleaner} & Photoguard  &  25.3614&  0.06022&  21.7390&  \textbf{0.19018}&  0.00939&  0.75646\\
                         & Glaze  &  25.7053&  0.03406&  22.8134&  0.14303&  0.00768&  0.78250\\
                         & AdvDM  &  24.6748&  0.04763&  22.4885&  0.16196&  0.00926&  0.75834\\
                         & SDS  &  \textbf{24.1779}&  0.05513&  22.1588&  0.16882&  0.01001&  0.74709\\
                         \rowcolor[gray]{0.9}     \cellcolor{white} & DIA-PT (Ours) &  25.3572&  0.03936&  21.9392&  0.18543&  0.00971&  0.75839\\
                         \rowcolor[gray]{0.9}     \cellcolor{white} & DIA-R (Ours) &  24.6166&  \textbf{0.06166}&  \textbf{20.6104}&  0.18917&  \textbf{0.01714}&  \textbf{0.73489}\\

 \cmidrule{1-8}
\multirow{6}{*}{Gaussian Noising} & Photoguard  &  26.1351&  0.0426&  21.5181&  \textbf{0.2908}&  0.0094&  0.5805\\
                         & Glaze  &  26.1265&  0.0364&  22.1301&  0.2591&  0.0084&  0.6048\\
                         & AdvDM  &  25.5851&  \textbf{0.0528}&  21.5729&  0.2772&  0.0104&  0.5920\\
                         & SDS  &  \textbf{25.3543}&  0.0515&  21.7844&  0.2720&  0.0101&  0.6031\\
                        \rowcolor[gray]{0.9}     \cellcolor{white}  & DIA-PT (Ours) &  26.2993&  0.0423&  21.4033&  0.2809&  0.0099&  \textbf{0.5788}\\
                         \rowcolor[gray]{0.9}     \cellcolor{white} & DIA-R (Ours)  &  25.9461&  0.0449&  \textbf{21.1744}&  0.2750&  \textbf{0.0115}&  0.5918\\

 \cmidrule{1-8}
\multirow{6}{*}{Noisy Upscaling} & Photoguard  &  25.4812&  \textbf{0.0381}&  23.0208&  0.1561&  0.0076&  0.7654\\
                         & Glaze  &  \textbf{25.4772}&  0.0351&  22.9662&  0.1506&  0.0077&  0.7658\\
                         & AdvDM  &  25.5246&  0.0344&  22.8962&  0.1582&  0.0076&  0.7553\\
                         & SDS  &  25.5639&  0.0355&  22.8581&  \textbf{0.1609}&  \textbf{0.0079}&  \textbf{0.7516}\\
                        \rowcolor[gray]{0.9}     \cellcolor{white}  & DIA-PT (Ours) &  25.5119&  0.0374&  \textbf{22.8195}&  0.1568&  0.0078&  0.7615\\
                         \rowcolor[gray]{0.9}     \cellcolor{white} & DIA-R (Ours)  &  25.6470&  0.0351&  22.8716&  0.1511&  \textbf{0.0079}&  0.7662\\
                         
\bottomrule
\end{tabular}
\caption{
Immunization performance across purification methods. This table demonstrates the robustness of various immunization methods against JPEG Compression, Crop \& Resize, and Adverse Cleaner attacks, evaluated on 700 images from the PIE-Bench dataset.}
\label{tab:purification}
}
\end{table*}

\newpage

\subsection{Considerations for Selecting Hyperparameters}

We provide an analysis of the hyperparameters of DIA-PT and DIA-R: attack iteration and trajectory length. Attack iteration is the number of PGD updates needed for optimization, while trajectory length is the length of the differentiable trajectory used in DDIM inversion and sampling during a single update.

Through Table~\ref{tab:attack_iteration}, we noted that both DIA-PT and DIA-R converge in disruption performance with just 20 attack iterations, which is likely because we precisely target the chained trajectory. Additionally, Table~\ref{tab:trajectory_length} reveals a difference between DIA-PT and DIA-R, with their best values found at trajectory lengths of 10 and 20, respectively. This indicates that for DIA-PT, trajectories beyond a certain length may have a negative impact since its loss is calculated based on the latent code $z_0$. Instead, DIA-R's performance improves with more detailed trajectories as it computes loss through $x_0$. To ensure a consistent inversion trajectory environment across all our experiments, we set the trajectory length to 10.


\begin{table*}[ht]
\fontsize{9}{11}\selectfont
\centering

\setlength{\tabcolsep}{4.6mm}{
\begin{tabular}{c|c|cccccc}
\toprule
              \textbf{Method}&        \textbf{Attack Iteration}& \textbf{CLIP}$\downarrow$ & \textbf{Distance $\uparrow$} & \textbf{PSNR $\downarrow$} & \textbf{LPIPS $\uparrow$} & \textbf{MSE $\uparrow$} & \textbf{SSIM $\downarrow$}  \\
\toprule
\multirow{4}{*}{DIA-PT} 
                    & 5  &  25.6048&  0.0482&  21.5865&  0.2094&  0.0103&  0.6992\\
                    & 10  &  24.3525&  0.0751&  20.0086&  0.2693&  0.0155&  0.6366\\
                    & 15  &  23.7979&  0.0913&  19.2879&  0.2949&  0.0188&  0.6078\\
                    & 20  &  \textbf{23.4575}&  \textbf{0.1006}&  \textbf{18.7744}&  \textbf{0.3124}&  \textbf{0.0208}&  \textbf{0.5874}\\
\cmidrule{1-8}
\multirow{4}{*}{DIA-R} 

                    & 5  &  24.6790&  0.0547&  20.8372&  0.1791&  0.0133&  0.7274\\
                    & 10   &  24.3205&  0.0670&  19.9336&  0.2038&  0.0186&  0.6967\\
                    & 15   &  23.8511&  0.0796&  19.3068&  0.2190&  0.0239&  0.6818\\
                    & 20   &  \textbf{23.4670}&  \textbf{0.0882}&  \textbf{18.7633}&  \textbf{0.2307}&  \textbf{0.0288}&  \textbf{0.6666}\\
\bottomrule
\end{tabular}
\caption{Attack Performance Across Different Attack Iterations. This table shows the performance of DIA-PT and DIA-R attacks using 5, 10, 15, and 20 attack iterations on the PIE-Bench dataset. The bold values represent the best performance across different attack iterations for each method.}
\label{tab:attack_iteration}
}
\end{table*}

\begin{table*}[ht]
\fontsize{9}{11}\selectfont
\centering
\setlength{\tabcolsep}{4.4mm}{
\begin{tabular}{c|c|cccccc}
\toprule
                     \textbf{Method}&        \textbf{Trajectory Length}& \textbf{CLIP}$\downarrow$ & \textbf{Distance $\uparrow$} & \textbf{PSNR $\downarrow$} & \textbf{LPIPS $\uparrow$} & \textbf{MSE $\uparrow$} & \textbf{SSIM $\downarrow$}  \\
\toprule
\multirow{3}{*}{DIA-PT}  & 5  &  25.6181&  0.0506&  21.1142&  0.2163&  0.0107&  0.6967\\
                        & 10  &  \textbf{23.4575}&  \textbf{0.1006}&  \textbf{18.7744}&  \textbf{0.3124}&  \textbf{0.0208}&  \textbf{0.5874}\\
                        & 20  &  24.1361&  0.0782&  20.0423&  0.2835&  0.0154&  0.6209\\

\cmidrule{1-8}

\multirow{3}{*}{DIA-R} 
                    & 5  &  24.3258&  0.0676&  19.7006&  0.2118&  0.0179&  0.6899\\
                    & 10  &  23.4670&  0.0882&  18.7633&  0.2307&  0.0288&  0.6666\\
                    & 20  &  \textbf{22.0941}&  \textbf{0.1101}&  \textbf{17.5972}&  \textbf{0.2540}&  \textbf{0.0432}&  \textbf{0.6451}\\
                    
\bottomrule
\end{tabular}
\caption{Attack Performance Across Different Trajectory Steps. This table shows the performance of DIA-R and DIA-PT attacks using 5, 10, and 20 trajectory steps on the PIE-Bench dataset. The bold values represent the best performance across different trajectory lengths for each method.}
\label{tab:trajectory_length}
}
\end{table*}

\newpage
\section{Transferability to Black-Box Models}\label{sec:suppl_black_box}
The Diffusion model is constantly updated and has an active developer community, resulting in many variants. As a result, the model used for attacks and the model used for editing the disrupted images may differ, potentially leading to attack performance degradation. This concept is referred to as model transferability, which indicates how well the disrupting performance is maintained across different scenarios. 
We conducted an experiment to test whether images disrupted using the initial stable diffusion model, version 1.4 (SD v1.4), retain their resistance when edited with black-box models, specifically stable diffusion versions 2.0 (SD v2.0) and 2.1 (SD v2.1). The experiment utilized a simple editing method DDIM-to-DDIM, and the hyperparameters used for editing were identical to those employed with SD v1.4, with the experiment conducted on the PIE-Bench.

\begin{table*}[ht!]
\fontsize{9}{11}\selectfont
\centering
\setlength{\tabcolsep}{4.2mm}{
\begin{tabular}{c|c|cccccc}
\toprule

                         \textbf{Diffusion Ver.}&        \textbf{Method}& \textbf{CLIP}$\downarrow$ & \textbf{Distance $\uparrow$} & \textbf{PSNR $\downarrow$} & \textbf{LPIPS $\uparrow$} & \textbf{MSE $\uparrow$} & \textbf{SSIM $\downarrow$}  \\ 
\toprule
\multirow{3}{*}{SD v1.4} & Natural Edit  &  25.7100&  0.02613&  23.8400&  0.09933&  0.00639&  0.80723\\
                         & \cellcolor[gray]{0.9}DIA-PT (Ours)&  \cellcolor[gray]{0.9}\textbf{23.4613} &  \cellcolor[gray]{0.9}\textbf{0.10042}&  \cellcolor[gray]{0.9}18.7803&  \cellcolor[gray]{0.9}\textbf{0.31218}&  \cellcolor[gray]{0.9}0.02074&  \cellcolor[gray]{0.9}\textbf{0.58770}\\
                         & \cellcolor[gray]{0.9}DIA-R (Ours) &  \cellcolor[gray]{0.9}23.4626&  \cellcolor[gray]{0.9}0.08821&  \cellcolor[gray]{0.9}\textbf{18.7655}&  \cellcolor[gray]{0.9}0.23087&  \cellcolor[gray]{0.9}\textbf{0.02877}&  \cellcolor[gray]{0.9}0.66656\\
                         
 \cmidrule{1-8}
\multirow{3}{*}{SD v2.0} & Natural Edit  &  25.7983&  0.04129&  23.2952&  0.12510&  0.00708&  0.79470\\
                         & \cellcolor[gray]{0.9}DIA-PT (Ours) &  \cellcolor[gray]{0.9}25.1747&  \cellcolor[gray]{0.9}0.05941&  \cellcolor[gray]{0.9}21.0996&  \cellcolor[gray]{0.9}0.22989&  \cellcolor[gray]{0.9}0.01152&  \cellcolor[gray]{0.9}0.70174\\
                         & \cellcolor[gray]{0.9}DIA-R (Ours)  &  \cellcolor[gray]{0.9}\textbf{24.1616}&  \cellcolor[gray]{0.9}\textbf{0.07028}&  \cellcolor[gray]{0.9}\textbf{20.1148}&  \cellcolor[gray]{0.9}\textbf{0.20655}&  \cellcolor[gray]{0.9}\textbf{0.01826}&  \cellcolor[gray]{0.9}\textbf{0.71447}\\
                         
\cmidrule{1-8}
\multirow{3}{*}{SD v2.1} & Natural Edit  &  24.5758&  0.05082&  21.7974&  0.16001&  0.01027&  0.76361\\
                         & \cellcolor[gray]{0.9}DIA-PT (Ours) &  \cellcolor[gray]{0.9}23.4423&  \cellcolor[gray]{0.9}\textbf{0.08422}&  \cellcolor[gray]{0.9}19.2938&  \cellcolor[gray]{0.9}\textbf{0.26466}&  \cellcolor[gray]{0.9}0.01734&  \cellcolor[gray]{0.9}\textbf{0.65941}\\
                         & \cellcolor[gray]{0.9}DIA-R (Ours)  &  \cellcolor[gray]{0.9}\textbf{22.5146}&  \cellcolor[gray]{0.9}0.08270&  \cellcolor[gray]{0.9}\textbf{19.1470}&  \cellcolor[gray]{0.9}0.23906&  \cellcolor[gray]{0.9}\textbf{0.02164}&  \cellcolor[gray]{0.9}0.68034\\
                         
\bottomrule
\end{tabular}
\caption{Disrupting Performance Comparison Across Stable Diffusion Model Versions. The table illustrates the robustness of images disrupted using the early version of Stable Diffusion (SD v1.4) when attempting editing attempts using different versions of the model (SD v2.0 and SD v2.1). The provided metrics (CLIP, Distance, PSNR, LPIPS, MSE, SSIM) evaluate various aspects of the edited images, showing that the immunized retain some immunity despite the difference in model versions. The arrow next to each metric name indicates the direction of better performance.}\label{tab:model_trans}
}
\end{table*}
\FloatBarrier 
In Table~\ref{tab:model_trans}, we observe that images immunized with the early version of Stable Diffusion (SD v1.4) retain a substantial disruptive signal when edited with different versions of Stable Diffusion. These results are crucial, as SD v2.1 and SD v2.0, along with the earlier SD v1.4, serve as the foundational models for most community-driven developments. 

\begin{figure}[h!]
\centering
\includegraphics[width=0.7\columnwidth]{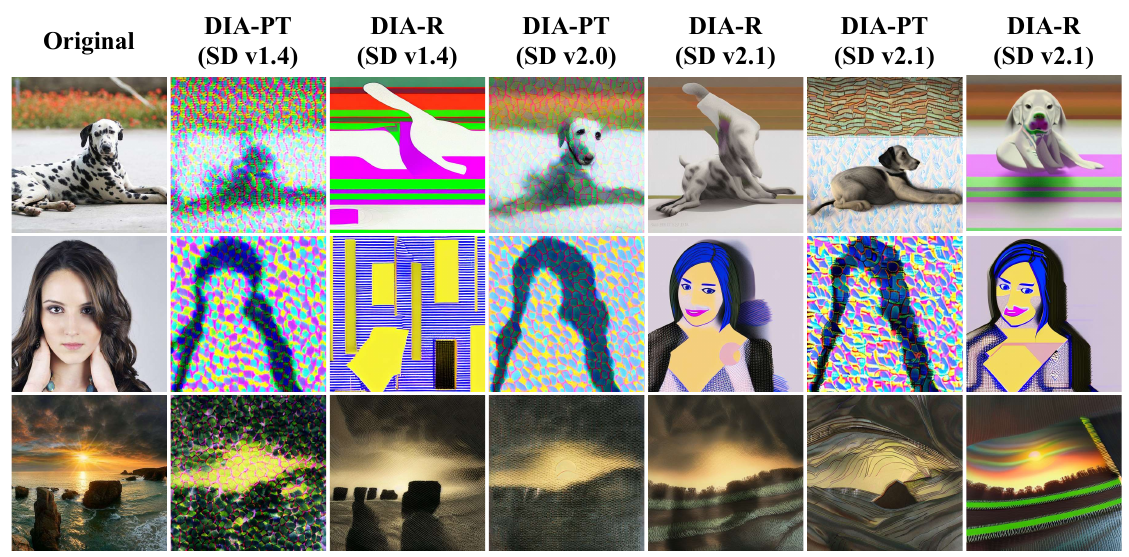}
\caption{
Quality comparison across Stable Diffusion Model Version. In this figure, DIA-PT and DIA-R visualize the results of editing images immunized in SD v1.4 across SD v1.4, SD v2.0, and SD v2.1. Editing in different versions reduces the disruptive performance, but still shows considerable effectiveness.
}
\label{fig:transferability}
\vspace{-8pt}
\end{figure}

Interestingly, our experiments consistently show that the attack is less disruptive when using different versions of Stable Diffusion (SD v2.0 and SD v2.1), but it remains a consistent disruption. Additionally, the qualitative result in Fig.~\ref{fig:transferability} enables visual understanding. Overall, the results support the generalizability of our approach, demonstrating that even with advancements in model versions, the disrupted images continue to exhibit strong resistance to editing attempts. 

\clearpage
\section{Observation on Over-Editing Scenarios}
\label{sec:ddim_p2p}

\begin{figure}[h!]
\centering
\includegraphics[width=\columnwidth]{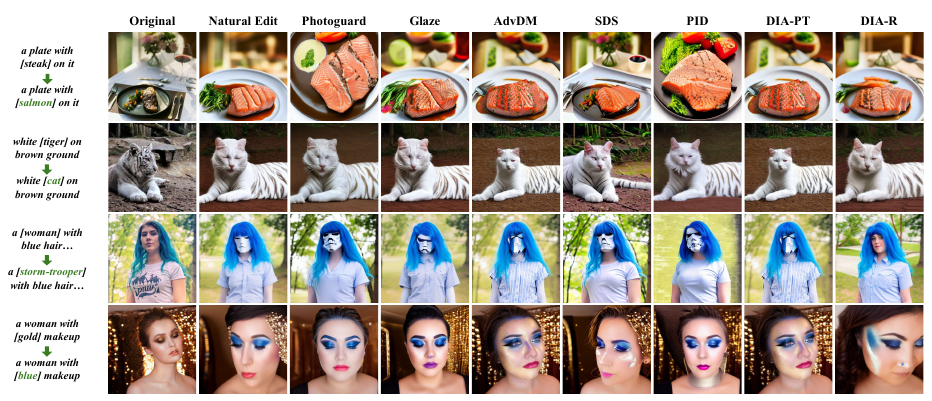}
\caption{
Quality comparison of images generated by DDIM-to-P2P across different immunization methods. The words in green indicate the parts to be edited from the original image. We visualize the failure to preserve the integrity of the original image.
}
\label{fig:ddim_p2p}
\vspace{-8pt}
\end{figure}

\begin{figure}[h!]
\centering
\includegraphics[width=\columnwidth]{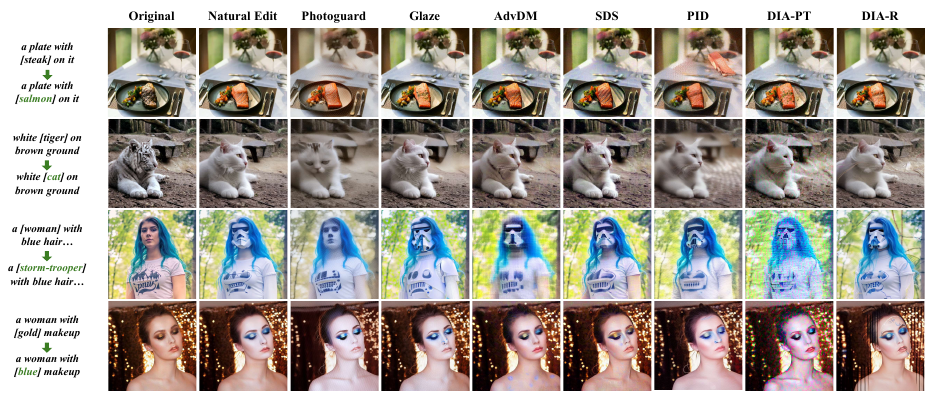}
\caption{
Quality comparison of images generated by Direct-to-P2P across different immunization methods. The words in green indicate the parts to be edited from the original image. We visualize that Direct-to-P2P robustly edits against immunization methods.
}
\label{fig:direct_p2p}
\vspace{-8pt}
\end{figure}

We extensively report over-edited images observed in DDIM-to-P2P and Direct-to-P2P. In Fig.~\ref{fig:ddim_p2p}, DDIM-to-P2P produces text-familiar images through P2P's aggressive attention map handling, which causes the failure to preserve the integrity of the original image during editing. In Fig.~\ref{fig:direct_p2p}, Direct-to-P2P shows a similar performance to Natural Edit as it corrects the target diffusion trajectory.

\clearpage
\section{Limitation}
Our proposed DIA-PT takes approximately 40 seconds, while DIA-R takes around 1 minute and 50 seconds. Although the required VRAM of 6-7GB is not overly demanding, there is room for improvement. Additionally, our method focuses on current image inversion methods and prominent image generation models. Should future image inversion methods evolve with operations orthogonal to the current DDIM inversion process, or the image modeling paradigm is subjected to changes, our method may undergo performance decay. We believe that analyzing our approach to address these limitations will help guide future research on the problem of image editing immunity through disruption.